\documentclass[letterpaper]{article} 
\usepackage{aaai25}  
\usepackage{times}  
\usepackage{helvet}  
\usepackage{courier}  
\usepackage[hyphens]{url}  
\usepackage{graphicx} 
\usepackage{natbib}  
\usepackage{caption} 
\usepackage{algorithm}
\usepackage{algorithmic}

\usepackage{newfloat}
\usepackage{listings}
\DeclareCaptionStyle{ruled}{labelfont=normalfont,labelsep=colon,strut=off} 
\floatstyle{ruled}
\newfloat{listing}{tb}{lst}{}
\floatname{listing}{Listing}
\usepackage{amsmath}
\usepackage[capitalize]{cleveref}
\crefname{section}{Sec.}{Secs.}
\Crefname{section}{Section}{Sections}
\Crefname{table}{Table}{Tables}
\crefname{table}{Tab.}{Tabs.}
\usepackage{tabularx}
\usepackage{booktabs}
\usepackage{amssymb}
\usepackage{bbding}
\usepackage{pifont}
\usepackage{wasysym}
\usepackage{utfsym}
\usepackage{fontawesome}
\usepackage{array}
\usepackage{subfigure}
\usepackage{multirow}
\usepackage{makecell}
\usepackage{dsfont}

\title{Combating Semantic Contamination in Learning with Label Noise}
\author {
    Wenxiao Fan,
    Kan Li\textsuperscript{\footnotemark[2]}
}
\affiliations {
    School of Computer Science, Beijing Institute of Technology\\
    \{wenxiaofan, likan\}@bit.edu.cn
}

\usepackage{bibentry}

\begin{document}

\maketitle

\renewcommand{\thefootnote}{\fnsymbol{footnote}}
\footnotetext[2]{Corresponding author.} 

\begin{abstract}
Noisy labels can negatively impact the performance of deep neural networks. One common solution is label refurbishment, which involves reconstructing noisy labels through predictions and distributions. However, these methods may introduce problematic semantic associations, a phenomenon that we identify as Semantic Contamination. Through an analysis of Robust LR, a representative label refurbishment method, we found that utilizing the logits of views for refurbishment does not adequately balance the semantic information of individual classes. Conversely, using the logits of models fails to maintain consistent semantic relationships across models, which explains why label refurbishment methods frequently encounter issues related to Semantic Contamination. To address this issue, we propose a novel method called Collaborative Cross Learning, which utilizes semi-supervised learning on refurbished labels to extract appropriate semantic associations from embeddings across views and models. Experimental results show that our method outperforms existing approaches on both synthetic and real-world noisy datasets, effectively mitigating the impact of label noise and Semantic Contamination.
\end{abstract}

%
\begin{links}
    \link{Extended version}{https://arxiv.org/abs/2412.11620}
\end{links}

\section{Introduction}


In recent years, notable progress has been achieved in various fields through deep learning methodologies \cite{DBLP:journals/corr/abs-2004-10934,DBLP:conf/cvpr/MarriottR021}. The use of labeled datasets plays a pivotal role in achieving these notable outcomes. Nevertheless, as datasets continue to expand in size, the probability of encountering noisy labels also increases. Such corrupted knowledge can be assimilated by models, which consequently results in a noticeable decrease in their performance \cite{DBLP:conf/iclr/ZhangBHRV17,Arplt2017}. This occurrence naturally prompts an urgent inquiry into how deep learning continues to succeed despite the presence of label noise.

%


State-of-the-art methods in Learning with Noisy Labels (LwNL) have notably enhanced noise robustness through label refurbishment methods \cite{Malach2017,Song2019,Chen2020BeyondCA,Chen2021TwoWD}.
The core idea of label refurbishment methods is to transform problematic labels into new, informative labels for the model to learn.
Previous researches \cite{DBLP:conf/nips/TarvainenV17, Song2019, Chen2021TwoWD} have shown that label refurbishment methods face self-reinforcing errors and confirmation bias, which can hinder performance.

\begin{figure}[t]
\centering
\includegraphics[width=1\columnwidth]{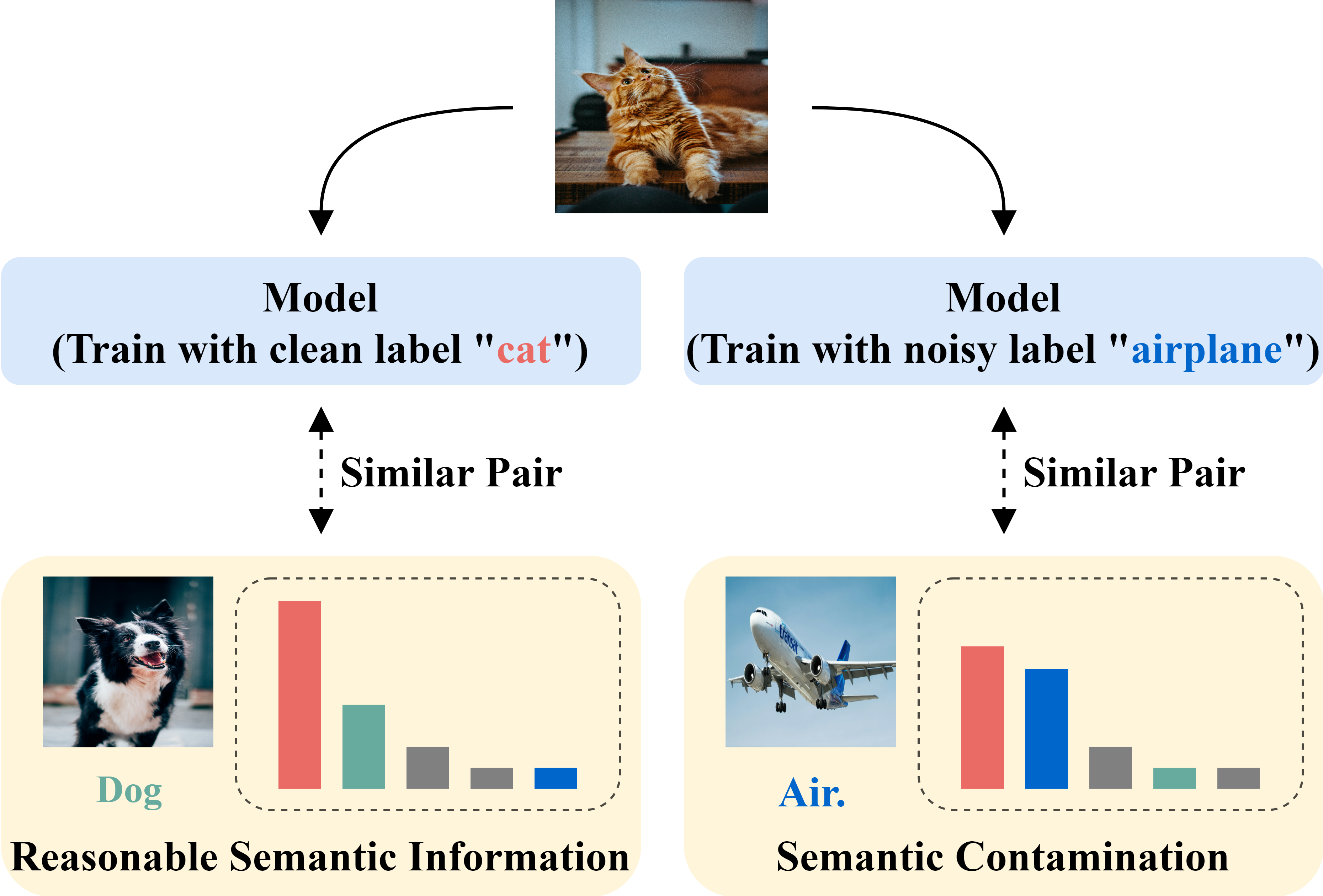} 
\caption{Illustration of Semantic Contamination. Air. is short for airplane. After being trained with noisy labels, models may learn the problematic semantic pairs, such as (cat, airplane) is more similar than (cat, dog). This issue may cause the model to learn incorrect feature spaces, impacting its performance. In this study, we mainly focus on how to enable the model to learn reasonable semantic information in order to overcome Semantic Contamination.}
\label{fig:intro_1}
\end{figure}

Meanwhile, we have also discovered that, in addition to the two previously mentioned drawbacks, label refurbishment methods face another significant issue that can be more detrimental to the model:  \textbf{Semantic Contamination} (SC). This scenario pertains to the model's inability to comprehend reasonable semantic associations, leading to the failure to acquire robust and consistent representations.
For instance, if a cat is mislabeled in an airplane, the similarity between a cat and a dog could be smaller than the similarity between a cat and an airplane, as illustrated in \cref{fig:intro_1}.
For clean datasets, reasonable semantic information can be easily learned by the model. However, in LwNL, samples in the same class could be clustered into different categories, leading to inconsistent predictions.
Hence, how to acquire the relevant semantic information remains an open question.

This study focuses on how label refurbishment methods can mitigate the impact of noisy labels and Semantic Contamination.
We initially analyze \textit{why the label refurbishment methods are prone to experiencing Semantic Contamination} and observe that directly aligning logits from different models, a common practice in label refurbishment, does not effectively align the embeddings. While aligning the logits of different views can cluster the embeddings, the unequal confidences for each class hinder the models from acquiring appropriate semantic information. Instead of relying solely on refurbishing with logits, we suggest mining latent relevancy in embeddings across views and models to learn semantic information and propose a novel method called Collaborative Cross Learning, which consists of two components: Cross-view learning and Cross-model learning. For Cross-view learning, we decouple the class label and the semantic concept and utilize self-supervised learning to prevent the incorporation of harmful semantic information. For Cross-model learning, we propose promoting the alignment of different models by using \textbf{C}ollaborative \textbf{C}ontrastive \textbf{L}earning \textbf{R}efurbished \textbf{L}abels (CCLRL) and theoretically establish that optimizing CCLRL enhances mutual information between the two models. The superiority of our method over state-of-the-art (SOTA) methods is demonstrated through validation on various synthetic and real-world benchmarks.
Our contributions can be summarized as follows:
\begin{itemize}
\item We introduce a new challenge called Semantic Contamination in LwNL and analyze the reason why label refurbishment methods may be susceptible to SC from the perspective of views and models.
\item We propose a novel method called Collaborative Cross Learning. By decoupling the semantic concept between views and mimicking the contrastive distributions between models, it successfully obtains robust and consistent representations while alleviating the damage of both Semantic Contamination and label noise.
\item Experimental results show that our method advances state-of-the-art results on CIFAR with synthetic label noise, as well as on real-world noisy datasets.
\end{itemize}


\section{Related Work}

\subsubsection{Label Refurbishment in LwNL.}
For label refurbishment, mainstream methods estimate refurbished labels through three ways:
1) Models: Decouple \cite{Malach2017} updates two models with only disagreed samples between models. SEAL \cite{Chen2020BeyondCA} retrains the model with the average predictions of the teacher model as refurbished labels.
2) Models and Views: RoLR \cite{Chen2021TwoWD} refurbishes noisy labels by aligning the predictions between models and views. 
3) Historical predictions: SELFIE \cite{Song2019} only includes samples with consistent predictions in recent epochs for refurbishment.
However, the process of refurbishing labels poses challenges in high-noise environments due to self-refining errors, which can impede the training of models.

\begin{figure}
    \centering
    \subfigure[T-SNE analyze for RoLR.]
     { \begin{minipage}[t]{0.45\linewidth}
			\centering
			\includegraphics[width=1\linewidth]{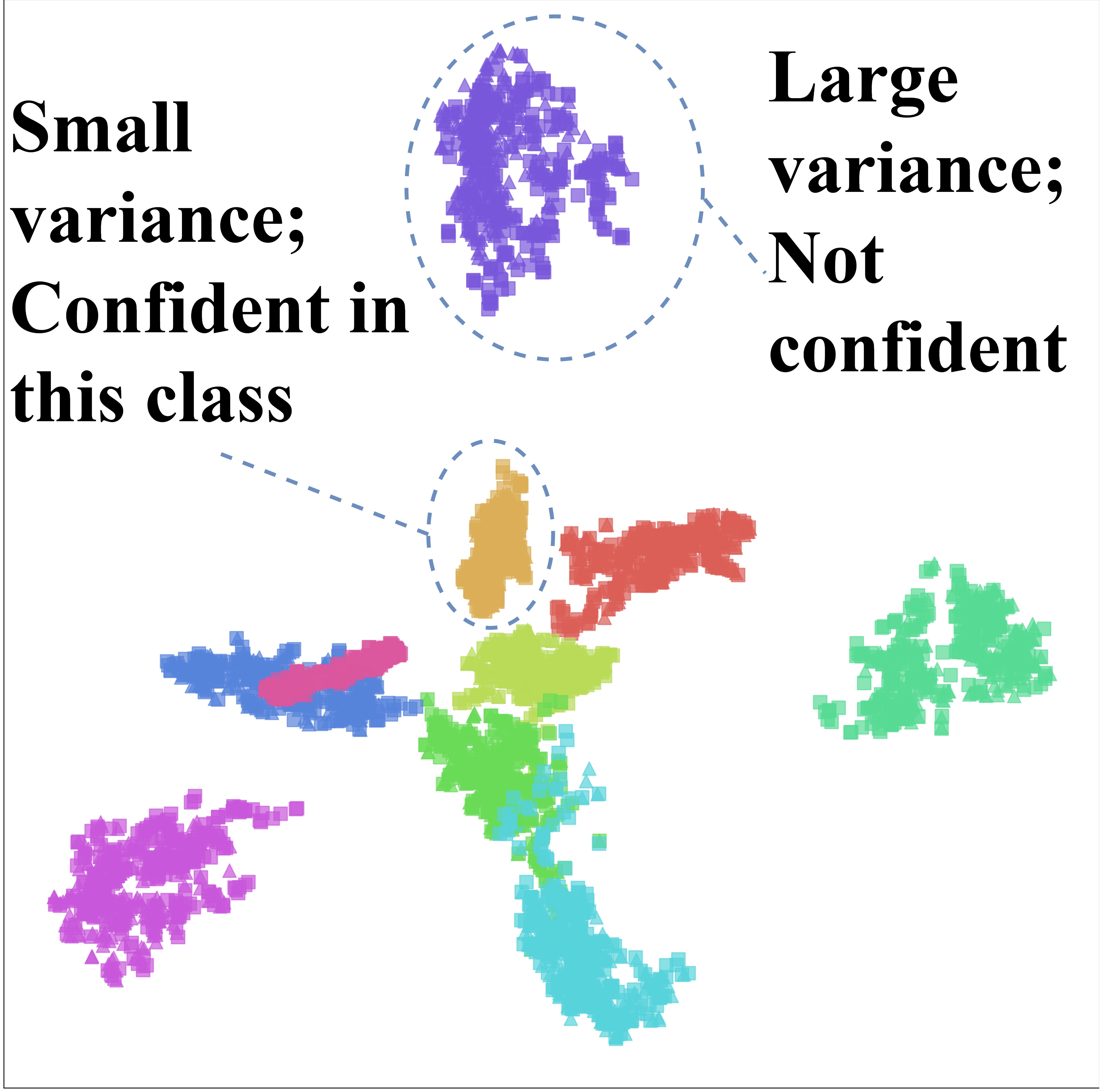}
			\label{fig:rolr_view_1}
		\end{minipage}
		}
    \hfill
    \subfigure[Entropy of variances of each class in CIFAR-10 under different symmetric noise (20\%, 50\% and 80\%).]
     { \begin{minipage}[t]{0.45\linewidth}
			\centering
			\includegraphics[width=1\linewidth]{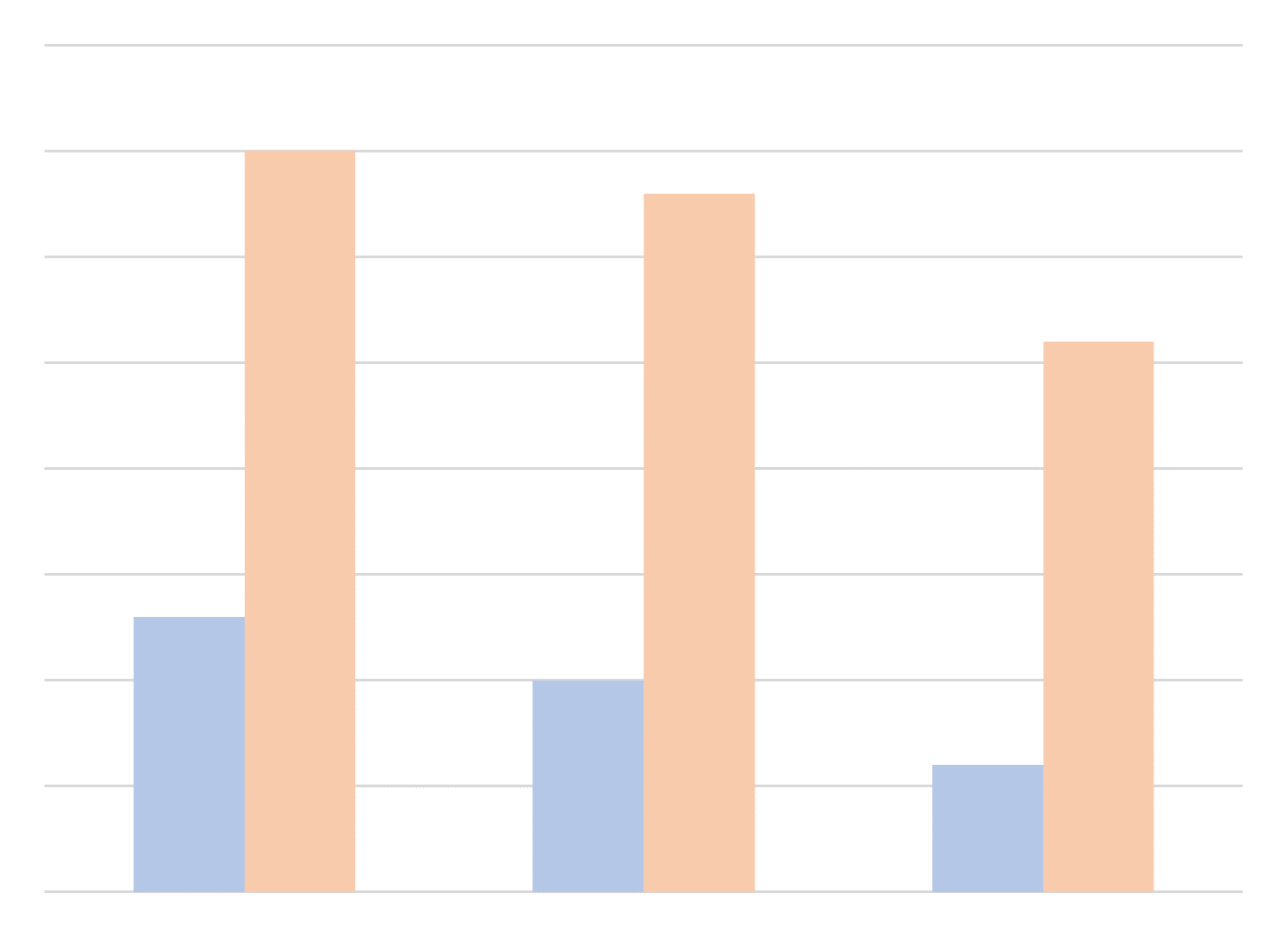}
			\label{fig:view_anly_1}
		\end{minipage}
		}
    \hfill
    \caption{Illustration of Semantic Imbalance Among Classes. \cref{fig:view_anly_1} shows that our method (orange) can learn more balance representations compared with RoLR (blue).}
    \label{fig:only_view}
\end{figure}

\subsubsection{Semi-supervised Learning.}
The field of semi-supervised learning (SSL) has experienced significant advances through the application of consistency regularization, which aims to minimize the disparity in model predictions between two views of the same sample or two predictions of the same sample using different models.
MixMatch \cite{DBLP:conf/nips/BerthelotCGPOR19} initially aligned outputs from different views and models, but the enhancement strategies for different views were consistent. In contrast, FixMatch \cite{DBLP:conf/nips/SohnBCZZRCKL20} utilized both strong and weak transformations, in addition to confidently pseudo-labeling, which led to favorable results. Currently, SSL is also implemented in LwNL, for instance, DivideMix \cite{DBLP:conf/iclr/LiSH20} incorporates MixMatch in LwNL, while RankMatch \cite{DBLP:conf/iccv/Zhang0FLCLL23} partially adopts FixMatch. RoLR \cite{Chen2021TwoWD} also employs distinct SSL augmentation strategies. However, SSL is susceptible to confirmation bias, and our experimental results indicate that SSL is still influenced by SC, both of which can impact the model's performance.

\section{Semantic Contamination in LwNL}
\label{sec:why}


As mentioned in the Introduction, Semantic Contamination refers to the phenomenon where, in the presence of latent semantic relationships among samples, a model fails to capture accurate semantic associations. The captured corrupted semantic information often corresponds to noisy labels.
Such toxic information is commonly utilized in various label refurbishment methods as part of the pseudo-labels and can harm generalization and lead to poor robustness in LwNL.
To address this issue, in this study, we first explore \textit{why label refurbishment methods are susceptible to Semantic Contamination}.
First, we take RoLR \cite{Chen2021TwoWD} as an example of label refurbishment methods and analyze the influence of different views and models through which RoLR may have learned potentially erroneous semantic information.
We demonstrate that relying solely on views for refurbishing  could lead to \textbf{semantic imbalance among classes}, ultimately impairing performance.
Apart from that, we find that relying on models for refurbishing cannot even maintain the \textbf{semantic consistency across models}. Addressing these drawbacks is the focus of our method.

\subsubsection{Preliminary.}
For convenience, notations within our work are clarified first. For a $C$-way image classification task with noisy labels, the training dataset is denoted by $D=\{({x},\hat{{y}})\}$, where $x$ is the training sample and $\hat{y}$ is the label which may be incorrectly annotated.
We denote the models in the training stage as $\theta=g(f(x))$, where $f$ is the feature extractor and $g$ represents the classifier. $p$ is the prediction and $p_i$ is the probability of $i$-th class of input distribution $p$.

RoLR, which we used as the example, is a SOTA method that integrates pseudo-labeling and confidence estimation techniques to refurbish noisy labels. In pseudo-labeling stage, the pseudo-labels $\tilde{y}$ is create by two averaged and sharpened models’ predictions from different augmentation strategies, as depicted in \cref{eq:rolr}. 

\begin{equation}
\begin{aligned}
    \tilde{y}^m = &\text{Sharpen}(\frac{p^w(y|x;\theta_m) + p^w(y|x;\theta_{(1-m)})}{2})\\
    &\text{Sharpen}(p)_i = \frac{p_i^{\frac{1}{T}}}{\sum_{j=1}^Cp_j^{\frac{1}{T}}}
    \label{eq:rolr}
\end{aligned}
\end{equation}
where $\tilde{y}^m$ is the pseudo-labels for model $\theta_m$, $m\in\{0,1\}$ means two models, $p^w,p^s$ mean weak and strong augmentation strategies for $p$. $C$ is the number of classes and $T$ is the temperature.
The loss of RoLR can be written by:
\begin{equation}
        \begin{aligned}
        \label{eq:rolr_loss}
        &\mathcal{L}(\mathbf{p}^s_{\theta_m},y^*) = \mathcal{L}_{\text{c}}(\mathbf{p}^s_{\theta_m},\omega\hat{y}+(1-\omega)\tilde{y}^m)\\
        &=\underbrace{\omega\mathcal{L}_{\text{c}}(\mathbf{p}^s_{\theta_m},\hat{y})}_{\text{Correct guidance}} + \frac{1-\omega}{2}(\underbrace{\mathcal{L}_{\text{c}}(\mathbf{p}^s_{\theta_m},\mathbf{p}^w_{\theta_m})}_{\text{Cross-view learning}} + \underbrace{\mathcal{L}_{\text{c}}(\mathbf{p}^s_{\theta_m},\mathbf{p}^w_{{\theta_{v}}})}_{\text{Cross-model learning}})
    \end{aligned}
\end{equation}
where $y^*$ is the refurbished label, $\omega$ is the label confidence obtained by the confidence estimation stage and $\mathbf{p}^w_{\theta_m}$ is simplified form of $p^w(y|x;\theta_m)$, $\mathcal{L}_{\text{c}}$ is the cross-entropy, $v=1-m$ for short.
We omit the Sharpen (set $T=1$) and decompose the loss into three parts in \cref{eq:rolr_loss}: Correct guidance, Cross-view learning and Cross-model learning.
The last two terms directly impact the semantic information and may cause harmful drawbacks: semantic imbalance among classes and semantic inconsistency among, respectively.

\subsubsection{Semantic imbalance among classes.}
Cross-view learning in \cref{eq:rolr_loss} aligns different views of the same model. To evaluate its influence specifically on semantic information, we decouple it into two components: one related to the predicted class $y_{\text{pred}} = \max\limits_ip^s_i$ and the other related to semantic concepts in \cref{eq:view_dec}.

\begin{equation}
    \label{eq:view_dec}
    \mathcal{L}_{\text{c}}(p^s,p^w) = \underbrace{\mathcal{L}_{\text{c}}(p_{y_{\text{pred}}}^s,p^w_{y_{\text{pred}}})}_{\text{Prediction guidance}} + \underbrace{\sum\limits_{i\neq y_{\text{pred}}}\mathcal{L}_{\text{c}}(p_i^s,p_i^w)}_{\text{Semantic smoothing}}
\end{equation}

While Prediction guidance ensures alignment of views of the samples during training, Semantic smoothing aims to obtain consistent semantic representations, as illustrated in \cref{fig:rolr_view_1}. However, Semantic smoothing can be highly vulnerable to noisy labels. Upon evaluating the entropy of variances of each class in \cref{fig:view_anly_1}, it is observed that the variances are significantly imbalanced due to the fail of Semantic smoothing. This phenomenon, named semantic imbalance among classes, leads to the model showing overconfidence in certain classes. Such an imbalance may introduce discrepancies in representations of different classes, which ultimately results in performance degradation.

\begin{figure}
    \centering
   \subfigure[Analyse of  $\mathcal{M}_{\text{embed}}$]
     { \begin{minipage}[t]{0.42\linewidth}
			\centering
			\includegraphics[width=1\linewidth]{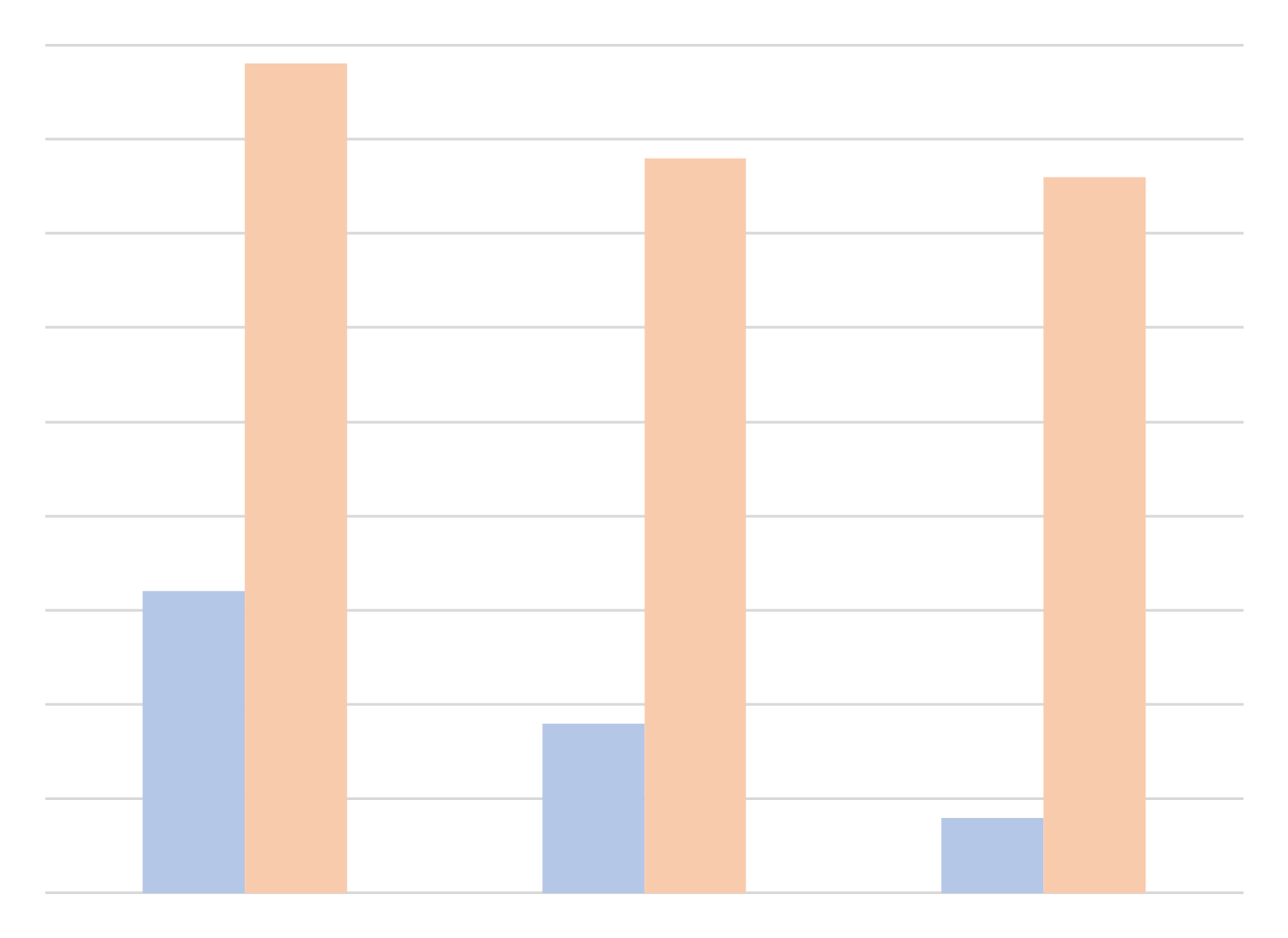}
			\label{fig:model_anly_1}
		\end{minipage}
		}
    \hfill
  \subfigure[Analyse of  $\mathcal{M}_{\text{logit}}$]
     { \begin{minipage}[t]{0.42\linewidth}
			\centering
			\includegraphics[width=1\linewidth]{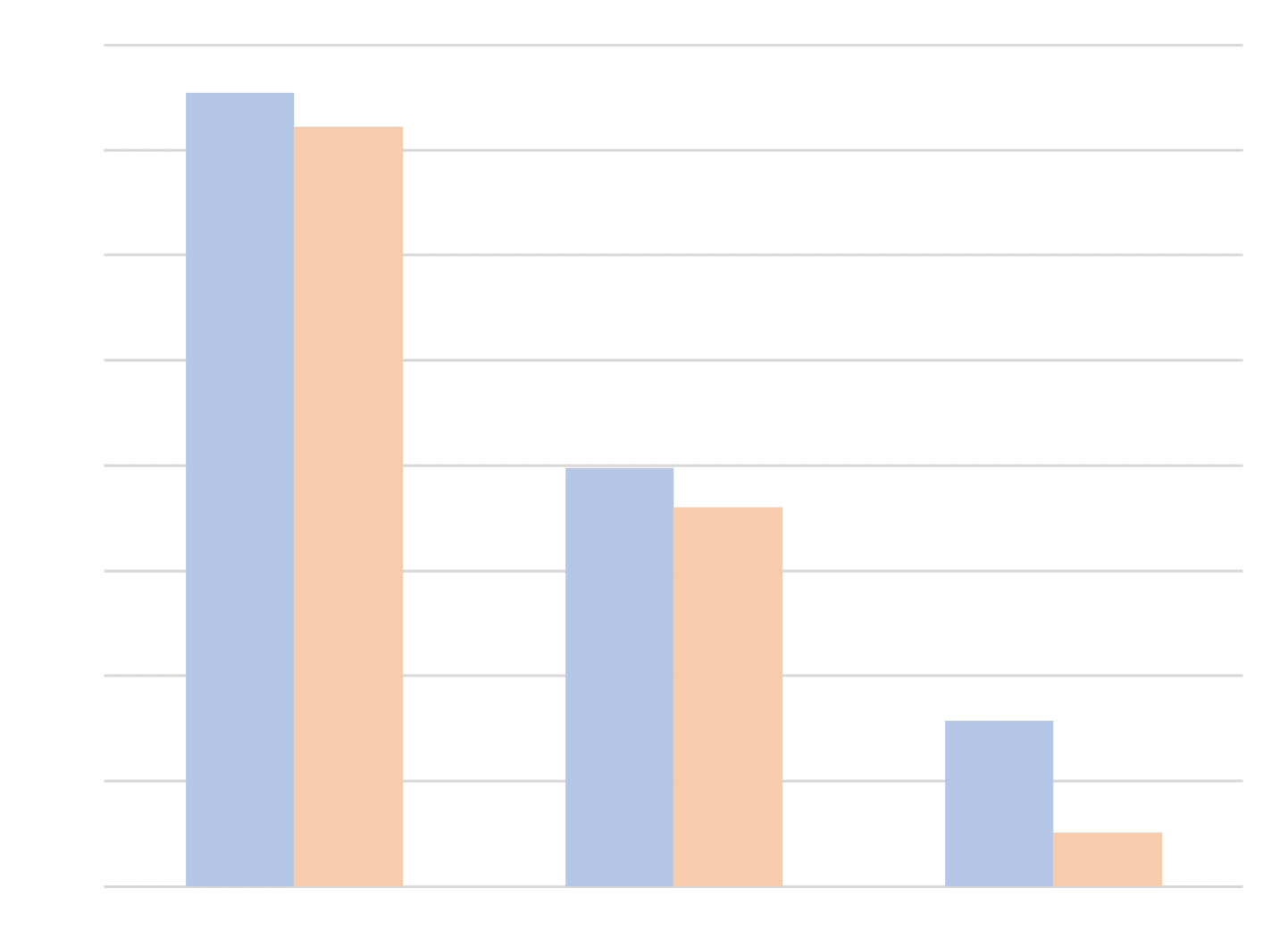}
			\label{fig:model_anly_2}
		\end{minipage}
		}
    \hfill
    \caption{Evaluation results of semantic consistency across models between RoLR (blue) with our method (orange) on CIFAR-10 with different symmetric noise (20\%, 50\% and 80\%).}
    \label{fig:model_anly}
\end{figure}

\subsubsection{Semantic inconsistency among models.}

Beside Cross-view learning in \cref{eq:rolr_loss}, Cross-model learning term offers an alternative way to gain semantic information across models by aligning the models in logits.  
In general, we aim to achieve semantic consistency among different models by learning a unified representation across models. However, the existence of noisy labels and the random initialization of different models impede the extraction of universal embeddings among models. Therefore, instead of pursuing this unattainable goal, our aim is to align the semantic relationships across models, ensuring semantic consistency across models.
To evaluate the alignment of semantic relationships, we investigate the following two metrics:
\begin{itemize}
    \item {Embedding-wise metric $\mathcal{M}_{\text{embed}}$: We utilize \cref{eq:metric_embed}  to assess the consistency of relationships between embeddings across models \cite{DBLP:journals/corr/abs-1301-3781}, such as $f_{\theta_1}(\text{dog}) = f_{\theta_1}(\text{cat}) - f_{\theta_0}(\text{cat}) + f_{\theta_0}(\text{dog})$. 
    \begin{equation}
    \begin{aligned}
        \label{eq:metric_embed}
        &\mathcal{M}_{\text{embed}}(D,\theta_0,\theta_1) = \\
        &\mathop{\mathbb{E}}\limits_{x_0,x_1\in D}[\mathcal{M}_{\text{cos}}(f_{\theta_0}(x_0)-f_{\theta_0}(x_1); f_{\theta_1}(x_0)-f_{\theta_1}(x_1))]
    \end{aligned}
    \end{equation}
    where $\mathcal{M}_{\text{cos}}$ is the cosine similarity. Larger $\mathcal{M}_{\text{embed}}$  shows more greater semantic consistency across models.
    }
    \item {Logits-wise metric $\mathcal{M}_{\text{logit}}$: The Wasserstein Distance of logits between two models can show the level of disparity among models. Models with larger disparities exhibit less semantic consistency. }
\end{itemize}

\cref{fig:model_anly} displays the two metrics for various methods. It is observed that both metrics for RoLR are weaker than those for our method, indicating that RoLR does not maintain semantic consistency across models, leading to models being unable to learn consistent and robust representations.

We conclude the above analysis:
\textit{Cross-view learning in label refurbishment leads to semantic imbalance among classes, while Cross-model learning results in semantic inconsistency across models, which are why label refurbishment methods fail to obtain the appropriate semantic information.}
This conclusion inspires us that simply refurbishing logits is vulnerable with noisy labels and insufficient for learning the right representations. It is necessary to incorporate independent modules that can extract reasonable semantic information, which we introduce next.

\section{Methodology}
The previous section concludes that refurbishments on logits are inadequate for accurate semantic information. To tackle this challenge, we propose a method, named Collaborative Cross Learning, that learns semantic relationships from embedding perspectives. Specifically, we propose two modules: Semantic-wise Decoupling with Confident Learning (SDCL), which can help in balancing the confidence among classes for Cross-view learning, and Embedding-based Interactive Alignment (EIA), which aids in aligning the models and maintaining semantic consistency across models for Cross-model learning.
The overall pipeline is shown in \cref{fig:temp_pipeline} and the algorithm pseudocode is in Appendix.


\textbf{Semantic-wise Decoupling with Confident Learning.}
Similar to \cref{eq:view_dec}, Semantic-wise Decoupling decouples the prediction into the predicted class and the semantic concepts.
To avoid the effect of noisy labels, instead of using the logits in Semantic smoothing term, we employ an \textbf{A}ugmentation-wise \textbf{C}ontrastive \textbf{L}earning approach on embeddings. Strong augmentations are used as anchors to align with weak augmentations to reduce training difficulty, as depicted in \cref{eq:SDCL_1}.

\begin{equation}
\begin{aligned}
  \label{eq:SDCL_1}
    \mathcal{L}_{\text{ACL}}(x,\theta)=& -\log q_{\theta}^{s\rightarrow w}\\
    =&-\log\frac{\exp(f_{\theta}^s(x)\cdot f_{\theta}^w(x))}{\sum^N_{j=1}\exp(f_{\theta}^s(x)\cdot f_{\theta}^w(x_j))}  
\end{aligned}
\end{equation}
where $N$ is the batch number, $q_{\theta}^{s\rightarrow w}$ is the contrastive distribution of the model $\theta$ from strong augmentations $s$ to weak augmentations $w$. 

In addition to directly aligning the representations of different views, we also use \textbf{KL} divergence to align the latent relationships of samples between views, named as \textbf{V}iew-wise \textbf{M}imicry:

\begin{equation}
  \label{eq:SDCL_2}
    \mathcal{L}_{\text{VM}}(x,\theta) = \text{\textbf{KL}}(q_{\theta}^{s\rightarrow w}||q_{\theta}^{w\rightarrow s})
\end{equation}

\begin{figure}
    \centering
    \includegraphics[width=1\linewidth]{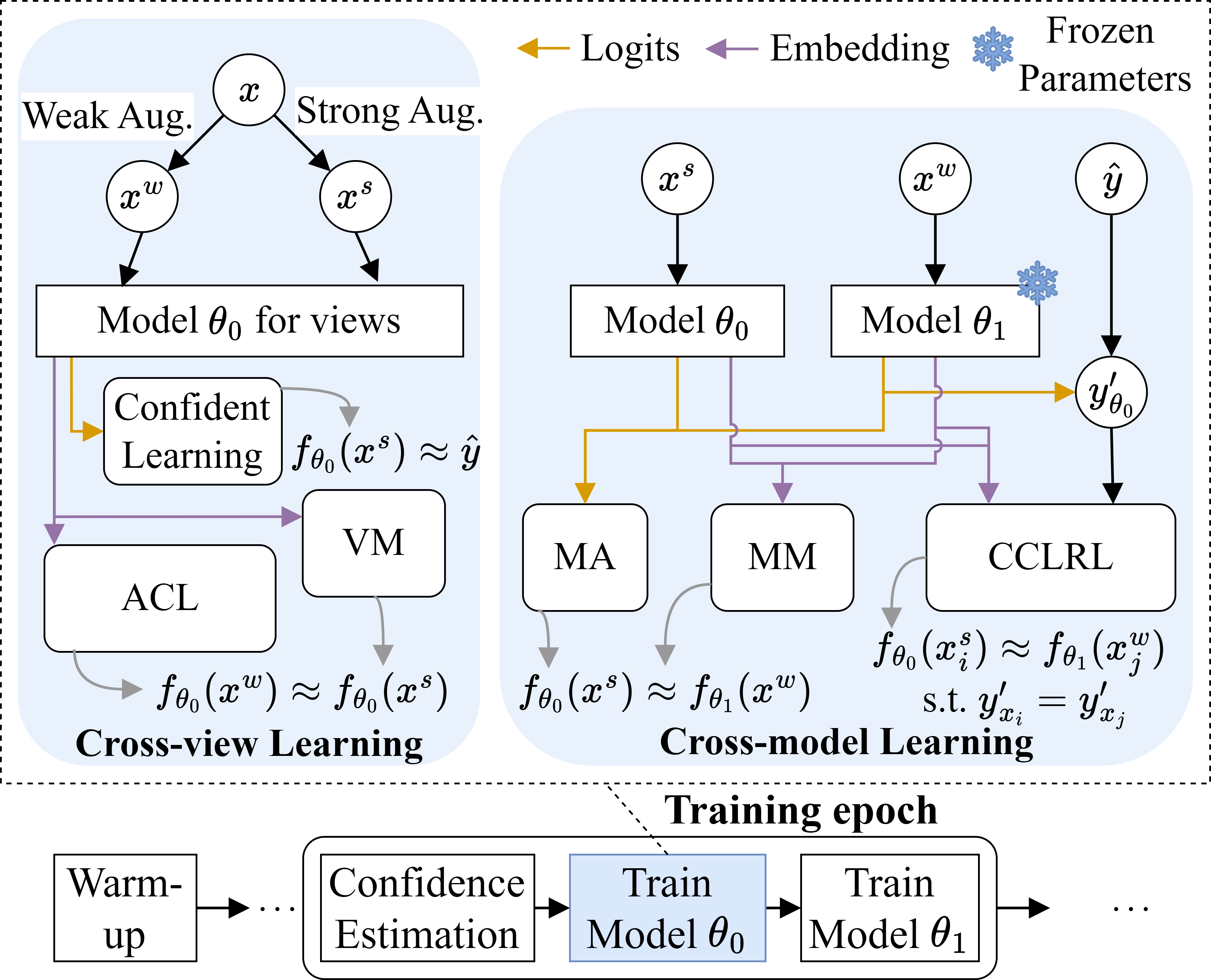}
    \caption{Pipeline of our method. The details of Warm-up and Confidence estimation by small-loss criterion and GMM can be found in Appendix.}
    \label{fig:temp_pipeline}
\end{figure}

For \textbf{P}redicted \textbf{G}uidance term in \cref{eq:view_dec}, considering the influence of noise that the prediction may be corrupted, the sample should only be learned when it exhibits a high confidence in the prediction as Confident Learning:
\begin{equation}
      \label{eq:SDCL_3}
      \mathcal{L}_{\text{PG}}(x,\theta) = -\mathds{1}[y_{\text{pred}} \geq c] 
      \cdot y_{\text{pred}}\log \mathbf{p}^s_\theta
\end{equation}
where $c$ is the threshold for confidence and $\mathds{1}[\cdot]$ is the characteristic function.

The loss of Cross-view Learning (CVL) can be calculated as follows:
\begin{equation}
    \mathcal{L}_{\text{CVL}} = \mathcal{L}_{\text{PG}} + \mathcal{L}_{\text{ACL}} + \mathcal{L}_{\text{VM}}
\end{equation}
$\mathcal{L}_{\text{PG}}$ is related to the predicted class and $\mathcal{L}_{\text{ACL}} + \mathcal{L}_{\text{VM}}$ can acquire relevant semantic information, separately. 

\textbf{Embedding-based Interactive Alignment.}
In order to get semantic consistency across models, in addition to aligning the logits in \cref{eq:rolr}, it is necessary to explore cross-model relationships in embeddings.
To address this, we introduce a novel approach called \textbf{C}ollaborative \textbf{C}ontrastive \textbf{L}earning on \textbf{R}efurbished \textbf{L}abels (CCLRL) to fully exploit the information interaction among diverse peer models. 
Specifically, CCLRL considers same-class samples with strong transformations $f_{\theta_m}^s(x)$ from $\theta_m$ and weak transformations $f_{\theta_{(1-m)}}^w(x)$ from $\theta_{(1-m)}$ as positive sample pairs, while treating all other samples pairs as negative sample pairs, as shown in \cref{eq:CCLRL_1}. 

\begin{equation}
\begin{aligned}
  \label{eq:CCLRL_1}
    &\mathcal{L}_{\text{CCLRL}}(x,\theta_m,\theta_{(1-m)})= -\log q_{\theta_m\rightarrow \theta_{(1-m)}}^{s\rightarrow w}\\
    &=-\log\frac{\sum\limits^N_{j=1,y'_{x} = {y'}_{x_j}}\exp(f_{\theta_m}^s(x)\cdot f_{\theta_{(1-m)}}^w(x_j))}{\sum\limits^N_{k=1}\exp(f_{\theta_m}^s(x)\cdot f_{\theta_{(1-m)}}^w(x_k))}  
\end{aligned}
\end{equation}

\begin{table*}[htbp]
  \centering
    \begin{tabular}{l|ccc|c|c|ccc|c|c}
    \toprule
    Dataset &      \multicolumn{5}{c|}{CIFAR-10} & \multicolumn{5}{c}{CIFAR-100} \\
    \midrule
    Noise Type & \multicolumn{3}{c|}{Sym}  & \multicolumn{1}{c|}{Pair} &  \multicolumn{1}{c|}{Ins}  & \multicolumn{3}{c|}{Sym}  & \multicolumn{1}{c|}{Pair}  & \multicolumn{1}{c}{Ins}   \\
    \midrule
    Method / Noise Rate & 20\% & 50\%& \multicolumn{1}{c|}{80\%}& \multicolumn{1}{c|}{40\%}  & \multicolumn{1}{c|}{40\%}  & 20\% & 50\%& \multicolumn{1}{c|}{80\%}& \multicolumn{1}{c|}{40\%} & \multicolumn{1}{c}{40\%}    \\
    \midrule
    Co-teaching \cite{Han2018c} & 88.2  & 50.7  & 21.1 & 55.3 & 59.5  & 58.5  & 33.0  & 5.8 & 39.2 & 40.7 \\
    SCE \cite{DBLP:conf/iccv/0001MCLY019}   & 89.9  & 78.5  & 31.8 & 58.8 & 77.8  & 55.9  & 40.2  & 12.9 & 39.9 &  42.4 \\
     

    JoCoR \cite{DBLP:conf/cvpr/WeiFC020}  & 89.4  & 53.3  & 25.8  &56.1 & 60.9  & 55.4  & 32.7  & 6.6 & 34.1 & 34.9 \\
    DivideMix \cite{DBLP:conf/iclr/LiSH20}& 95.7 &	94.4&	92.9	&	92.1	&	95.1& 76.9	&74.2&	59.6&	52.3&	76.1\\
    Co-learning \cite{DBLP:conf/mm/TanXWL21} & 91.8  & 79.3  & 37.0 & 66.3 &  78.9  & 70.3  & 63.9  & 38.9 & 49.1 & 62.9  \\
     SELC+ \cite{Lu2022SELCSL} &{{94.9}}& 87.2 &{78.6} & 88.1 & 84.2 &{76.4}& 62.4 &{37.2} & 45.2 & 44.3 \\

    RoLR \cite{Chen2021TwoWD} & 96.4&	95.7&	94.2&	92.8 & 	93.7&  78.6&	74.6&	66.2& 76.1 & 77.2  \\
    RankMatch \cite{DBLP:conf/iccv/Zhang0FLCLL23} & 96.4 &	95.4&	94.2&  94.4 & 93.8  & 79.3&	77.6&	67.2&  75.8& 76.5  \\
    CrossSplit \cite{DBLP:conf/icml/KimBZL23} & 96.9	&96.3	&95.4			&96.0 & 95.8 &79.9& 75.7 & 64.6&76.8 & 79.2  \\
    DISC \cite{DBLP:conf/cvpr/Li0S023} & 96.3 &95.4	&92.9&94.6	&	96.0 & 78.6	&76.3&	59.3&75.1  &78.4 \\
    DMLP (Naive) \cite{DBLP:conf/cvpr/TuZLLLWWZ23}& 94.2&	94.0 &93.2&93.9& 93.2& 72.3& 70.1	&	63.2& 71.8 & 72.2\\
    DMLP (DivideMix) \cite{DBLP:conf/cvpr/TuZLLLWWZ23} &96.2	&95.6	&94.3&	95.0  &95.4 & 79.4	&76.1&	68.5& 76.4 &  78.9 \\
    \midrule
    Ours &  \textbf{97.0}&	\textbf{96.5}&	\textbf{94.6}&  \textbf{96.1}	&	\textbf{96.2} & \textbf{79.5}	&\textbf{77.4}&	\textbf{70.3} & \textbf{77.2}	 &	\textbf{80.0} \\
    \bottomrule
    \end{tabular}%
    \caption{Comparison with state-of-the-art methods on CIFAR-10/100 datasets under various types of noise. 
    The results of other methods are from the published results of corresponding papers.
    The best results are indicated in bold.
     }
  \label{tab:main_1}%
\end{table*}%

Given the presence of noisy labels, the identification of samples of the same class is based on refurbished collaborative labels $y'$. 
Furthermore, to avoid self-refining errors, the collaborative refurbished labels for the current model $\theta_{m}$ are obtained from another model $\theta_{(1-m)}$, as shown in \cref{eq:CCLRL_2}.
\begin{equation}
    \label{eq:CCLRL_2}
    y'_{\theta_m}=\omega\hat{y}+(1-\omega)\mathbf{p}^w_{{\theta_{(1-m)}}}
\end{equation}

\textbf{Theoretical Analysis.} We attribute the effectiveness of minimizing \cref{eq:CCLRL_1} to maximizing the lower bound on the mutual information $I(f_{\theta_m}(x), f_{\theta_{(1-m)}}(x))$ between $\theta_m$ and $\theta_{(1-m)}$, which can be formulated as:
\begin{equation}
  \label{eq:CCLRL_3}
    I(f_{\theta_m}(x), f_{\theta_{q}}(x)) \geq \log(N)-\mathds{E}[\mathcal{L}_{\text{CCLRL}}(x,\theta_m,\theta_{v})]
\end{equation}

Inspired by \cite{DBLP:conf/eccv/TianKI20,DBLP:conf/aaai/YangACX22}, a detailed proof is provided in Appendix. Intuitively, the mutual information $I(f_{\theta_m}(x), f_{\theta_{(1-m)}}(x))$ measures the uncertainty in contrastive embeddings from $\theta_{(1-m)}$ when the anchor embeddings from $\theta_{m}$ are known \cite{DBLP:conf/aaai/YangACX22}. 
Higher mutual information implies that $\theta_m$ gains additional contrastive knowledge from others, ensuring the preservation of semantic consistency across models.

Similar to View-wise Mimicry, \textbf{M}odel-wise \textbf{M}imicry aligns the contrastive distribution $q$ of samples between models: 
\begin{equation}
  \label{eq:CCLRL_4}
    \mathcal{L}_{\text{MM}}(x,\theta_m,\theta_{(1-m)}) = \text{\textbf{KL}}(q_{\theta_m\rightarrow \theta_{(1-m)}}^{s\rightarrow w}||q_{\theta_{(1-m)}\rightarrow \theta_m}^{w\rightarrow s})
\end{equation}

Combining with the model alignment with logits in \cref{eq:rolr}, the loss for Cross-model learning (CML) can be formulated as:
\begin{equation}
    \mathcal{L}_{\text{CML}} = \mathcal{L}_{\text{c}}(\mathbf{p}^s_{\theta_m},\mathbf{p}^w_{{\theta_{(1-m)}}}) + \mathcal{L}_{\text{CCLRL}}
    + \mathcal{L}_{\text{MM}}
\end{equation}

\subsubsection{Overall Loss Function.}
Following \cite{DBLP:conf/iclr/LiSH20,Chen2021TwoWD,DBLP:conf/iccv/Zhang0FLCLL23}, we apply the regularization $\mathcal{L}_{\text{div}}$ to increase the diversity of predictions:
\begin{equation}
    \mathcal{L}_{\text{div}} = \sum\limits_{i=1}^C\frac{1}{C}\log(\frac{1}{C}/\frac{\sum_{j=1}^Np^s_{j}[i]}{N})
\end{equation}
where $p^s_{j}[i]$ is the $i$-th class of the prediction of the strong augmentation of sample $x_j$.

The overall loss function for model $\theta_m$ optimization is as follows:
\begin{equation}
\label{eq:all_loss}
    \mathcal{L} = \omega\mathcal{L}_{\text{c}}(\mathbf{p}^s_{\theta_m},\hat{y}) + \frac{1-\omega}{2}(\mathcal{L}_{\text{CVL}}+\mathcal{L}_{\text{CML}}) + \mathcal{L}_{\text{div}}
\end{equation}

Following \cite{DBLP:conf/iclr/LiSH20,Chen2021TwoWD}, we warm up the models and estimate the label confidence $\omega$ using the small-loss criterion and Gaussian Mixture Model (GMM). The detailed algorithm is provided in the Appendix.

\section{Experiments}
\subsection{Experimental Setup}

\subsubsection{Datasets.} 
To verify the effectiveness of our method, we perform our method on classification tasks with six benchmarks: CIFAR-10 \cite{krizhevsky2009learning}, CIFAR-100 \cite{krizhevsky2009learning}, CIFAR-10N \cite{DBLP:conf/iclr/WeiZ0L0022},  CIFAR-100N \cite{DBLP:conf/iclr/WeiZ0L0022}, Animal-10N \cite{Song2019} and WebVision \cite{DBLP:journals/corr/abs-1708-02862}. The last four benchmarks are real-world noisy datasets.

\begin{table*}[htbp]
  \centering
    \begin{tabular}{ll|ccccc|c}
    \toprule
    \multicolumn{2}{l}{Dataset}  \vline & \multicolumn{5}{c}{CIFAR-10N} \vline& \multicolumn{1}{l}{CIFAR-100N} \\
    \multirow{2}{*}{Method}  & Noise Type & Aggre & Rand1 &Rand2 &Rand3 & Worst & Fine   \\
          & Noise Rate & 9.0\% & 17.2\% &18.12\% &17.64\%& 40.2\% & 40.2\% \\
        \midrule
    
        \multicolumn{2}{l}{Decouple  \cite{Malach2017}}\vline & 88.1  & 85.5  & 85.5  & 85.4  & 74.6  & 45.5   \\
    \multicolumn{2}{l}{GCE  \cite{DBLP:conf/nips/ZhangS18}}\vline & 89.8  & 88.0  & 87.8  & 87.8  & 75.3  & 50.3    \\

    \multicolumn{2}{l}{Co-teaching  \cite{Han2018c}} \vline& 89.9  & 87.8  & 87.2  & 87.4  & 62.3  & 40.5   \\
    \multicolumn{2}{l|}{DivideMix \cite{DBLP:conf/iclr/LiSH20}}&93.2 & 92.8 &92.6 & 93.1 & 89.2 & 55.2  \\
    \multicolumn{2}{l}{Co-learning   \cite{DBLP:conf/mm/TanXWL21}}\vline & 92.4  & 91.3  & 91.2  & 91.4  & 81.0  & 47.9   \\
    Cores + LC  \cite{Wei2022MitigatingMO} & & 92.1 & 90.9& 91.2&91.1&{82.8}&{59.2} \\
     RoLR \cite{Chen2021TwoWD} & & 95.4 & 94.9 & 94.7 & 95.2 & 92.3 & 62.3 \\
    RankMatch \cite{DBLP:conf/iccv/Zhang0FLCLL23}& & 95.6 & 94.8 & 95.1 & 95.3 & 92.8 & 65.2 \\
    \midrule
    Ours & & \textbf{96.4 }&	\textbf{96.0}&	\textbf{95.8}	&\textbf{96.1}	&\textbf{93.1}	& \textbf{65.5}  \\
    \bottomrule
    \end{tabular}%
    \caption{Comparison with state-of-the-art methods on CIFAR-N. The results are from \cite{Wei2022MitigatingMO} and our replication.}
  \label{tab:main3}%
\end{table*}%

\subsubsection{Synthetic noise injection.}
\label{Sec:exp_setup}
Under such an assumption that the corruption process is conditionally independent of data features when the true label is given \cite{DBLP:journals/corr/abs-2007-08199,DBLP:conf/iclr/ZhangBHRV17},
we can construct the dataset containing noises by the noise transition matrix $T$ \cite{DBLP:journals/corr/abs-2007-08199,Jiang2017,Malach2017,Song2019}, 
where $T_{ij} \overset{\underset{\mathrm{def}}{}}{=} p( \hat{y}=j| y= i)$ is the probability of the clean label $i$ being corrupted into a noisy label $j$.
$T$ can model two types of noises:
(1) symmetric noise (Sym): $\forall i\neq j,T_{ij} = \frac{\tau_0}{C-1}$
and (2) pair noise (Pair): $\exists i\neq j,T_{ij} =\tau_0 \land \forall k\neq i,k\neq j,T_{ik} = 0 $, which is also known as the asymmetric noise. To evaluate the performance on varying noise rates from light noises to heavy noises, we run our method and other methods on different noise rates $\tau_0\in\{0.2, 0.4, 0.5, 0.8\}$.

\begin{table}[htbp]
    \centering
    \begin{tabular}{l|cc|cc}
    \toprule
      \multirow{2}{*}{Method}   & \multicolumn{2}{c|}{WebVision} & \multicolumn{2}{c}{ILSVRC12}  \\
         & top-1 & top-5 & top-1 & top-5  \\
         \midrule
         Co-teaching 	&63.6 & 85.2& 61.5& 84.7 \\
Iterative-CV &	65.2& 85.3& 61.6& 85.0 \\
DSOS &	77.8& 92.0& 74.4& 90.8 \\
DivideMix &	77.3& 91.6& 75.2& 90.8 \\
UNICON &	77.6& 93.4& 75.3& 93.7 \\
RoLR & 	81.8& 94.1& 75.5& 93.8 \\
RankMatch &	79.9 & 93.6 & 77.4 & 94.3 \\
         \midrule
         Ours & \textbf{82.3} & \textbf{94.6} & \textbf{78.2}  & \textbf{94.9}\\
         \bottomrule
    \end{tabular}
    \caption{Comparison with state-of-the-art methods on (mini) WebVision dataset. Numbers denote top-1 (top-5) accuracy (\%) on the WebVision and ImageNet ILSVRC12 validation set. The results of other methods are from \cite{Chen2021TwoWD,DBLP:conf/iccv/Zhang0FLCLL23}. }
    \label{tab:main_5}
\end{table}

We have also conducted experiments with another type of noise: instance-dependent noise (Ins) \cite{Chen2020BeyondCA}.
Unlike the above two types of noise, Ins allows the label noise to depend mandatorily on the samples, and optionally on the labels.

Weak and strong augmentations in our paper are identical with \cite{Chen2021TwoWD}. All the results from our runs are the average test accuracy over the last 10 epochs. We replicated experimental results that were missing from the corresponding papers.
A detailed description of Comparison Methods and Implementation Details can be found in Appendix.

\subsection{Experimental Results}

\subsubsection{Results on CIFAR with synthetic noise.}

\cref{tab:main_1} illuminates our method outperforms the state-of-the-art models across all noisy levels on CIFAR-10/100 with different types of synthetic noise.
Benefiting from the advanced semantic mining mechanism, our method can boost the averaged test accuracy of the last 10 epochs from 94.2\% to 94.6\% on CIFAR-10 and 66.2\% to 70.3\% on CIFAR-100 under the extreme case of 80\% noise compared with RoLR \cite{Chen2021TwoWD}. Compared with sample selection methods such as RankMatch \cite{DBLP:conf/iccv/Zhang0FLCLL23}, our method can achieve around 3.1\% (70.3\% v.s. 67.2\%) on CIFAR-100 with 80\% noisy labels.
We surpass DMLP \cite{DBLP:conf/cvpr/TuZLLLWWZ23}, a recent semi-supervised learning-based method, under all noise ratios, especially on the more challenging CIFAR-100 dataset with extreme noise.

\begin{table}[htbp]
    \centering
    \begin{tabular}{cccccc}
    \toprule
          SELFIE& PLC & NCT & RoLR & DISC  & Ours\\
         \midrule
         81.8 & 83.4 &  84.1& 88.5 & 87.1 & \textbf{89.7} \\
         \bottomrule
    \end{tabular}
    \caption{Comparison with other methods on Animal-10N. The results of other methods are from \cite{Chen2021TwoWD}.}
    \label{tab:main_4}
\end{table}

\subsubsection{Results on real-world datasets.}
\cref{tab:main3}, \cref{tab:main_5} and \cref{tab:main_4} demonstrate the results on CIFAR-N, Animal-10N and WebVision, respectively. 
Our method demonstrates superior performances compared to all other methods across various noisy real-world datasets.
In particular, compared with RoLR, our method achieves 1.2\% performance gains in Aniaml-10N.  Besides, compared with UNICON \cite{DBLP:conf/cvpr/KarimRRMS22}, another hybrid method that combines the advantages of semi-supervised learning and contrastive learning, our method surpasses the SOTA by over 3\% in top-1 accuracy on both mini-WebVision and ILSVRC12 validation sets, ensuring the SOTA top-5 accuracy on WebVision and ILSVRC12, demonstrating the effectiveness of our method.


\subsubsection{Results on Semi-supervised learning methods in LwNL.}
In addition to label refurbishment methods, recent research has widely adopted semi-supervised learning mechanisms like MixMatch \cite{DBLP:conf/nips/BerthelotCGPOR19}, such as DivideMix \cite{DBLP:conf/iclr/LiSH20}, to overcome the impact of noise. However, these methods also face the challenge of Semantic Contamination. To address this issue, we integrate $(\mathcal{L}_{\text{CVL}}$ and $\mathcal{L}_{\text{CML}})$ into the existing loss function to further enhance the learning of better semantic information.
From the results shown in \cref{tab:exp_ssl}, we can uncover the following empirical results: 1) Semi-supervised learning methods do not actually receive proper semantic information. By utilizing both $(\mathcal{L}_{\text{CVL}}$ and $\mathcal{L}_{\text{CML}})$, the model can acquire better semantic information, resulting in performance enhancement. 2) Ensuring semantic consistency between the two models ($\mathcal{L}_{\text{CML}}$) can yield greater performance improvements compared to learning semantic information between views ($\mathcal{L}_{\text{CVL}}$). This also implies the importance of addressing semantic inconsistencies between two models.

\begin{table}[htbp]
    \centering
    \begin{tabular}{l|cc|cc}
    \toprule
    Dataset & \multicolumn{2}{c|}{CIFAR-10} & \multicolumn{2}{c}{CIFAR-100} \\
    \midrule
        Noise ratio & 50\% & 80\% & 50\% & 80\% \\
        \midrule
        DivideMix & 94.4 &  92.9 & 74.2 & 59.6 \\
        \quad + $\mathcal{L}_{\text{CVL}}$ & 95.2 &  93.3 & 74.8 & 62.3 \\
        \quad + $\mathcal{L}_{\text{CML}}$ & 95.9 &  93.7 & 75.3 & 64.2 \\
        \quad + $(\mathcal{L}_{\text{CVL}} + \mathcal{L}_{\text{CML}})$ & \textbf{96.1} &  \textbf{94.1} & \textbf{76.1} & \textbf{68.4}  \\
        \bottomrule
    \end{tabular}
    \caption{Results on the combination of DivideMix and our method in terms of test accuracy (\%) on CIFAR-10 and CIFAR-100 with symmetric noise.}
    \label{tab:exp_ssl}
\end{table}

\subsubsection{Results for combating Semantic Contamination.}
To validate the effectiveness of our method in combating Semantic Contamination, we initially train our method and RoLR on CIFAR-100 with 80\% symmetric noise. We pick up samples with accurate predictions but differing second-largest logits and then identify samples from the class associated with the second-largest logits in the same batch and compute the similarity among these samples. As illustrated in \cref{tab:exp_csc}, we observe that: 1) Although the predictions are accurate, RoLR is still influenced by Semantic Contamination, which tends to assign higher similarity to pairs that lack semantic relevance (e.g. automobiles and cats). 2) in comparison to RoLR, our method assigns higher similarities to samples with semantic relationships (e.g. automobiles and trucks). This demonstrates that our method can capture more relevant semantic information, allowing the model to acquire a more continuous and consistent representation space.


\begin{table}
     \centering
    \begin{tabular}{ll|cccc}
        && \includegraphics[width=0.09\linewidth]{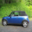} & \includegraphics[width=0.09\linewidth]{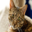} & \includegraphics[width=0.09\linewidth]{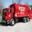} & \includegraphics[width=0.09\linewidth]{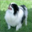}\\
        &Truth Label& Auto. & Cat & Truck & Dog \\
        &Noisy Label& Auto. & Ship & Horse & Dog \\
        \midrule
        \multirow{2}{*}{RoLR} & Prediction & Auto. & Ship & Horse & Dog \\
         &Similarity & - & 0.79 & 0.56 & 0.42 \\
         \midrule
      \multirow{2}{*}{Ours} & Prediction & Auto. & Ship & Horse & Dog  \\
       & Similarity & - & 0.52 & 0.78 & 0.49 \\
        \bottomrule
    \end{tabular}
    \caption{Results on Semantic Contamination on CIFAR-10 with 80\% symmetric  noise. Auto. is short for Automobile. Similarity refers to the cosine similarity between the embeddings of the first sample and the other three samples.}
    \label{tab:exp_csc}
\end{table}

\subsection{Ablation Study}

\subsubsection{Effects of components of our method.}
We remove the corresponding components to study the effects of each component of our method: 
We remove $\mathcal{L}_{\text{CVL}}$ and $\mathcal{L}_{\text{CML}}$ to validate the effect of these two modules, respectively. 
Moreover, we remove both the mimicry in \cref{eq:SDCL_2} and \cref{eq:CCLRL_4} to  validate the effect of the mimicry. 
We also remove the threshold in \cref{eq:SDCL_3} to validate the effect of the threshold for filtering the noise.
We replace the loss on the refurbished labels in \cref{eq:CCLRL_1} with the vanilla contrastive learning.
As shown in \cref{tab:exp_ab_compents}, the results validate the effectiveness of each component of our method. The mimicry, the threshold in $\mathcal{L}_{\text{PG}}$ and refurbished labels in \cref{eq:CCLRL_1} are beneficial to our method.

\begin{table}[htbp]
    \centering
    \begin{tabular}{l|cc|cc}
    \toprule
    Dataset & \multicolumn{2}{c|}{CIFAR-10} & \multicolumn{2}{c}{CIFAR-100} \\
    \midrule
        Noise ratio & 50\% & 80\% & 50\% & 80\% \\
        \midrule
        Ours & \textbf{96.5} & \textbf{94.6} & \textbf{77.4} & \textbf{70.3}  \\
        w/o Mimicry & 95.7 & 93.8 & 76.7 & 68.2\\
        w/o Threshold in $\mathcal{L}_{\text{PG}}$ & 95.8 & 92.3 & 75.2 & 67.9 \\
        w/o Refurbished labels & 96.1 & 93.2 & 76.2 & 68.4 \\
        w/o $\mathcal{L}_{\text{CVL}}$ & 95.2 & 92.4  & 76.5 & 68.2\\
        w/o $\mathcal{L}_{\text{CML}}$ & 94.8 & 91.4  & 75.1 & 66.2 \\
        \bottomrule
    \end{tabular}
    \caption{Ablation study results of test accuracy (\%) on CIFAR-10 and CIFAR-100 with symmetric  noise.}
    \label{tab:exp_ab_compents}
\end{table}


\subsubsection{The role of different augmentation strategies in LwNL.}

To explore the role of different augmentation strategies in LwNL, we focus on the effects of different data augmentation strategies on model results. Specifically, we directly train with the exception of different augmentation strategies, everything else remains the same.
In \cref{fig:exp_ds}, from left to right, the three data augmentation strategies are: both weak (blue), one weak and one strong (green) and both strong (orange), we find that using the same augmentation can affect performance. 
We speculate that the reason is that various augmentations can increase the diversity of training samples.



\subsubsection{Sensitivity Analysis.}
\begin{figure}
\begin{minipage}[tb]{0.32\linewidth}
     \centering
    \includegraphics[width=1\linewidth]{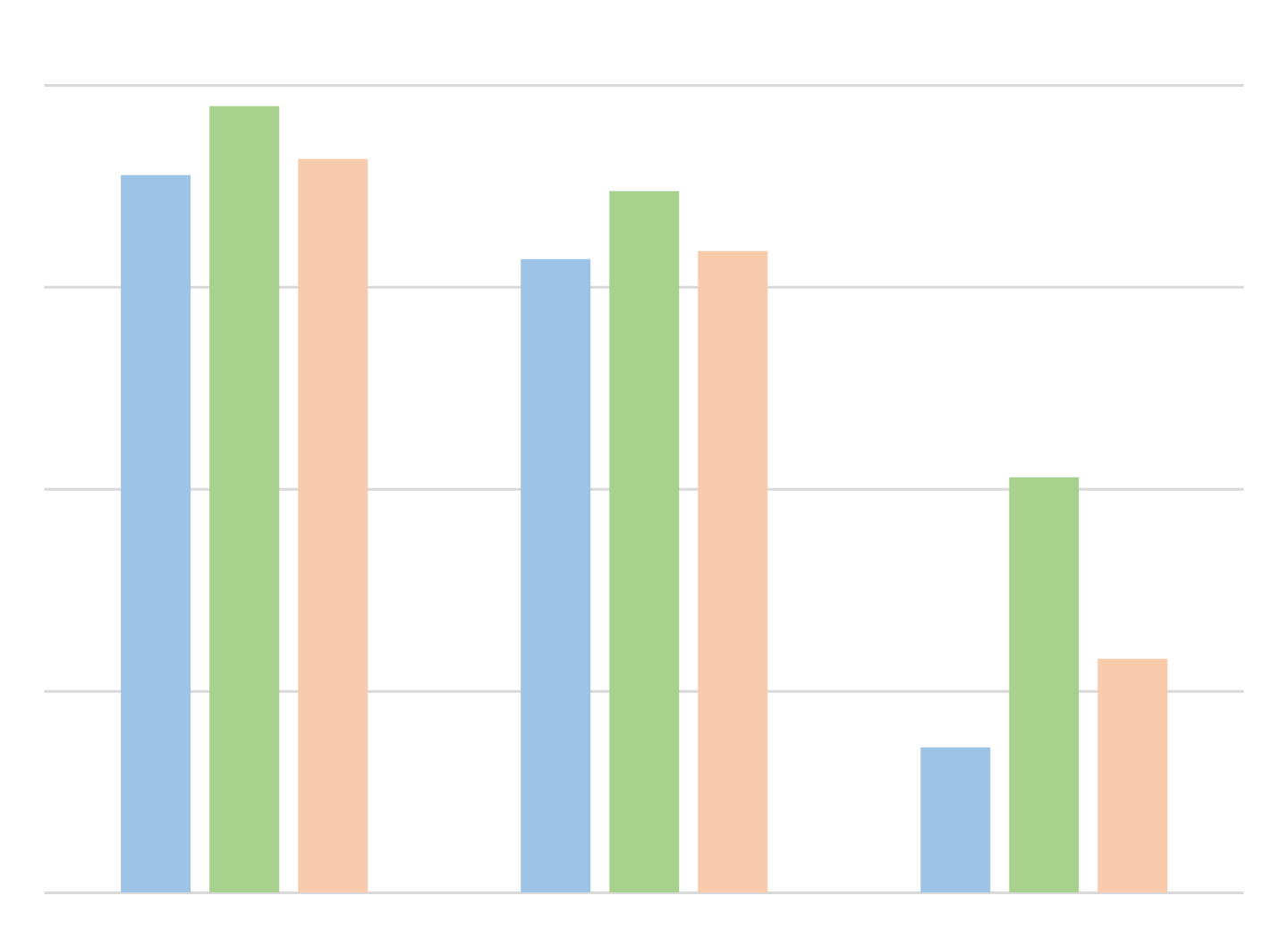}
    \caption{Results on different augmentation strategies on CIFAR-100 under 80\% symmetric noises.}
    \label{fig:exp_ds}
\end{minipage}
\hfill
    \begin{minipage}[tb]{0.65\linewidth}
    \centering
    \subfigure[Temperature $\tau$ from 0.05 to 0.9]
     { \begin{minipage}[t]{0.46\linewidth}
			\centering
			\includegraphics[width=1\linewidth]{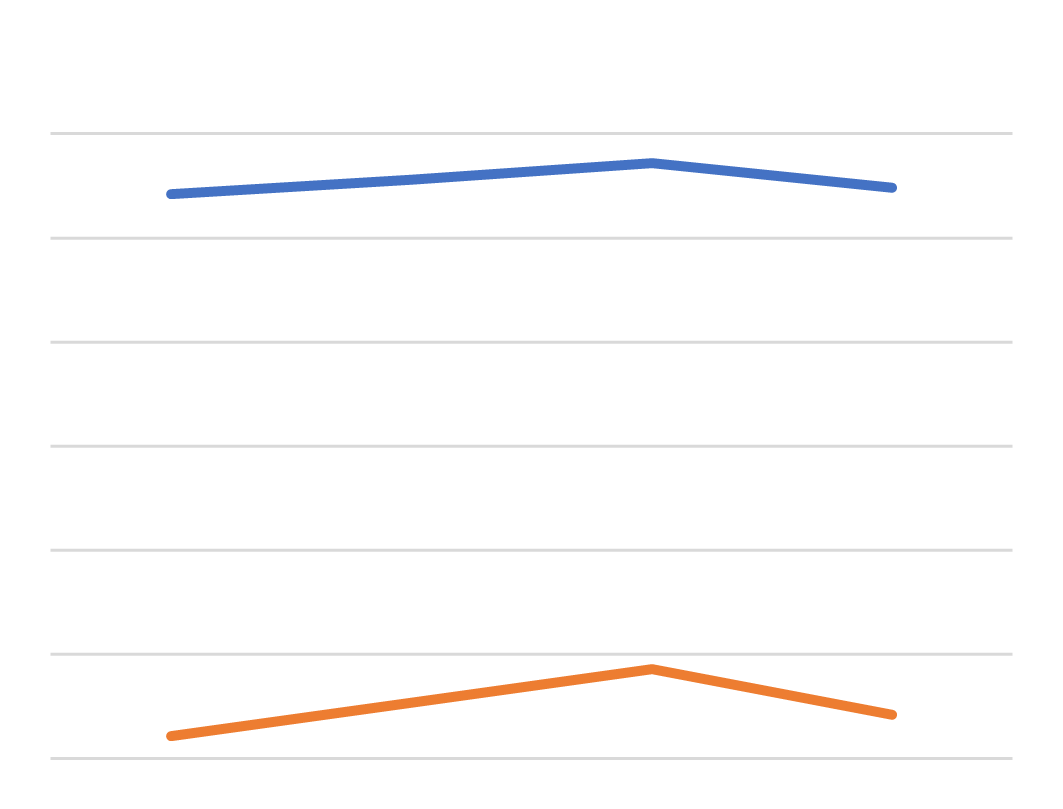}
			\label{fig:sen_tau}
		\end{minipage}
		}
    \subfigure[Threshold $c$ from 0.90 to 0.99]
     { \begin{minipage}[t]{0.46\linewidth}
			\centering
			\includegraphics[width=1\linewidth]{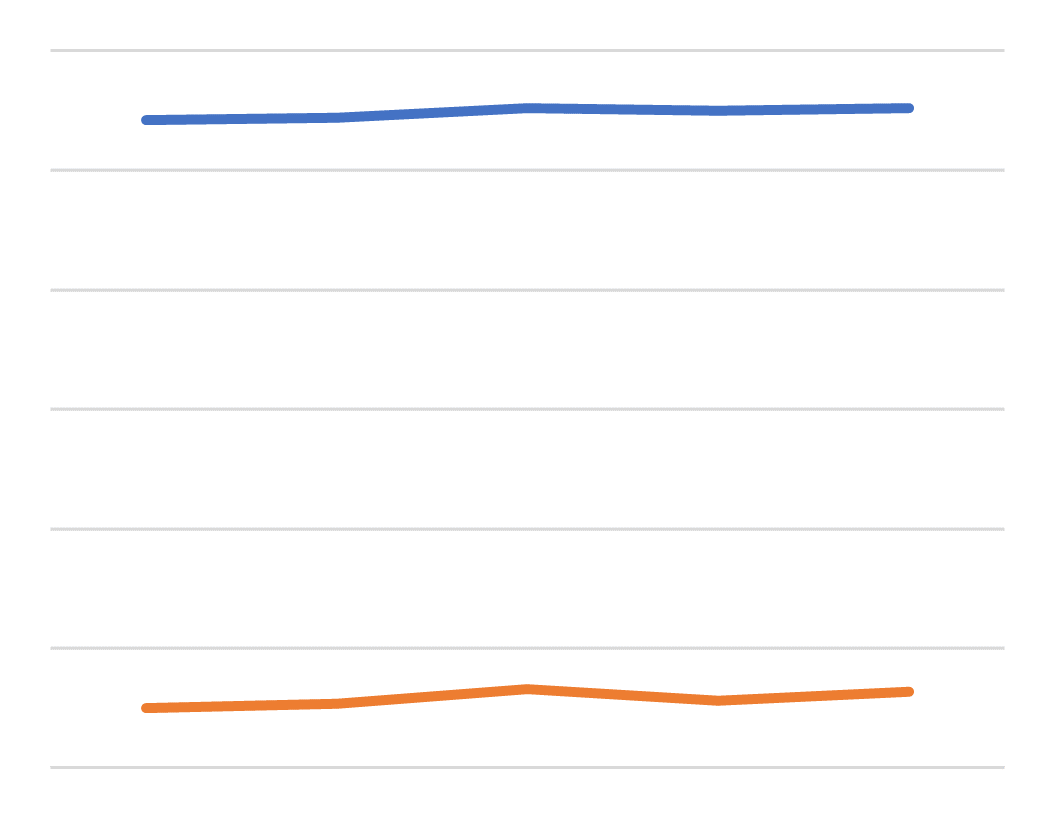}
			\label{fig:sen_c}
		\end{minipage}
		}
    \caption{Sensitivity to the variance of hyperparameters. Experiments are conducted on CIFAR-10 (blue) and CIFAR-100 (orange) under 80\% symmetric noises.}
    \label{fig:exp_plot}
\end{minipage}
\end{figure}

Our method introduces the temperature $\tau$ in \cref{eq:SDCL_1} and \cref{eq:CCLRL_1} and the threshold $c$ in \cref{eq:SDCL_3} as hyper-parameters.
We vary $\tau$ from 0.05 to 0.9 and range $c$ from 0.90 to 0.99. 
\cref{fig:sen_tau} shows that $\tau$ should be reasonable, as setting it too high or too low can both degrade the model's performance.
\cref{fig:sen_c} shows that our method is robust against various choices for $c$.


\section{Conclusion}
In this paper, we study the problem of Semantic Contamination together with label noise. We analyze the drawbacks of label refurbishment methods and explain why these methods cannot overcome Semantic Contamination. The conclusion motivates us to propose our method that can learn the more reasonable semantic information through Semantic Decoupling with Confident Learning and Emebdding-based Interactive Alignment. 
Experimental results illustrate that our method surpasses current methods in performance on synthetic and real-world noisy datasets, effectively reducing the influence of label noise and Semantic Contamination.
In the future, we will further explore the theoretical foundation and generalization analysis of our method.

\section{Acknowledgments}
This work is supported by Beijing Natural Science Foundation (No.4222037, L181010).

\bibliography{aaai25}

\clearpage







\newpage
\section{Appendix}

\subsection{Details of our method}

\subsubsection{Warm-up.}
\cite{Arplt2017} shows that the deep neural network tend to learn clean samples first. 
Therefore, we follow the process of recent work \cite{Han2018c,Song2019,Chen2021TwoWD} that warms two models in the early stages.  In the warm-up phase, the objective function is the vanilla cross-entropy without performing label refurbishment operations.
For CIFAR-like datasets, we consider the first 30 epochs as warm-up. For ImageNet-like datasets, we consider the first 80 epochs as warm-up.

\subsubsection{Confidence estimation.} 
Recent studies \cite{Arplt2017,Han2018c,Song2019,Chen2021TwoWD} have demonstrated that models are inclined to present smaller losses on clean samples. Therefore, we can use the loss value to judge whether the sample is clean or not.
Following the process in \cite{Han2018c,Song2019,Chen2021TwoWD}, we first calculate the cross-entropy loss per-sample between the noisy label and the prediction through the other model $\theta_{(1-m)}$ and use a two-component one-dimensional GMM to separate the datasets for training the current model $\theta_m$, as shwon in \cref{eq:ce}.

\begin{equation}
\label{eq:ce}
    \mathcal{W} = \text{GMM}(\{\mathcal{L}_i(\hat{y},\mathbf{p}^s_{\theta_{(1-m)}})\})
\end{equation}
where $\mathcal{W}=\{\omega_i\}$ is the label confidence which equals to the probability of each sample belonging to the GMM component with a smaller mean \cite{Chen2021TwoWD}.
The training of GMM follows the standard practice \cite{Song2019}.

\subsubsection{Pesudo-code of our method.}
As shown in \cref{alg:ccl}, we perform the pesudo-code of our method.

\begin{algorithm}[htb]
\caption{Collaborative Cross Learning}
\label{alg:ccl}
\textbf{Input}: $\mathit{E}$(epochs), $\mathit{W}$(Warm-Up epochs), $c$, $T$\\
\textbf{Output}: $\theta_0,\theta_1$ (model parameter) 

\begin{algorithmic}[1] 
 \STATE   Randomly initialize two models $\theta_0,\theta_1$.
 \FOR{$e \leftarrow 1$ to $E$} 
    \IF{$e<W$}
     \STATE   Train $\theta_0,\theta_1$ by Cross-entropy. \qquad$\triangleright$ Warm-up
    \ELSE
    \FOR{$m=0,1$}
    \STATE Obtain the label confidence $\mathcal{W}$. 
    \STATE \qquad\qquad\qquad\qquad$\triangleright$ Confidence estimation
    \STATE Get the augmentations for samples.
    \STATE Train the model $\theta_m$ using \cref{eq:all_loss}.
    \ENDFOR
    \ENDIF
 \ENDFOR
\STATE \textbf{return} $\theta$
\end{algorithmic}
\end{algorithm}


\subsection{T-SNE results on Semantic inconsistency across models}

\begin{figure}
    \centering
    \includegraphics[width=0.8\linewidth]{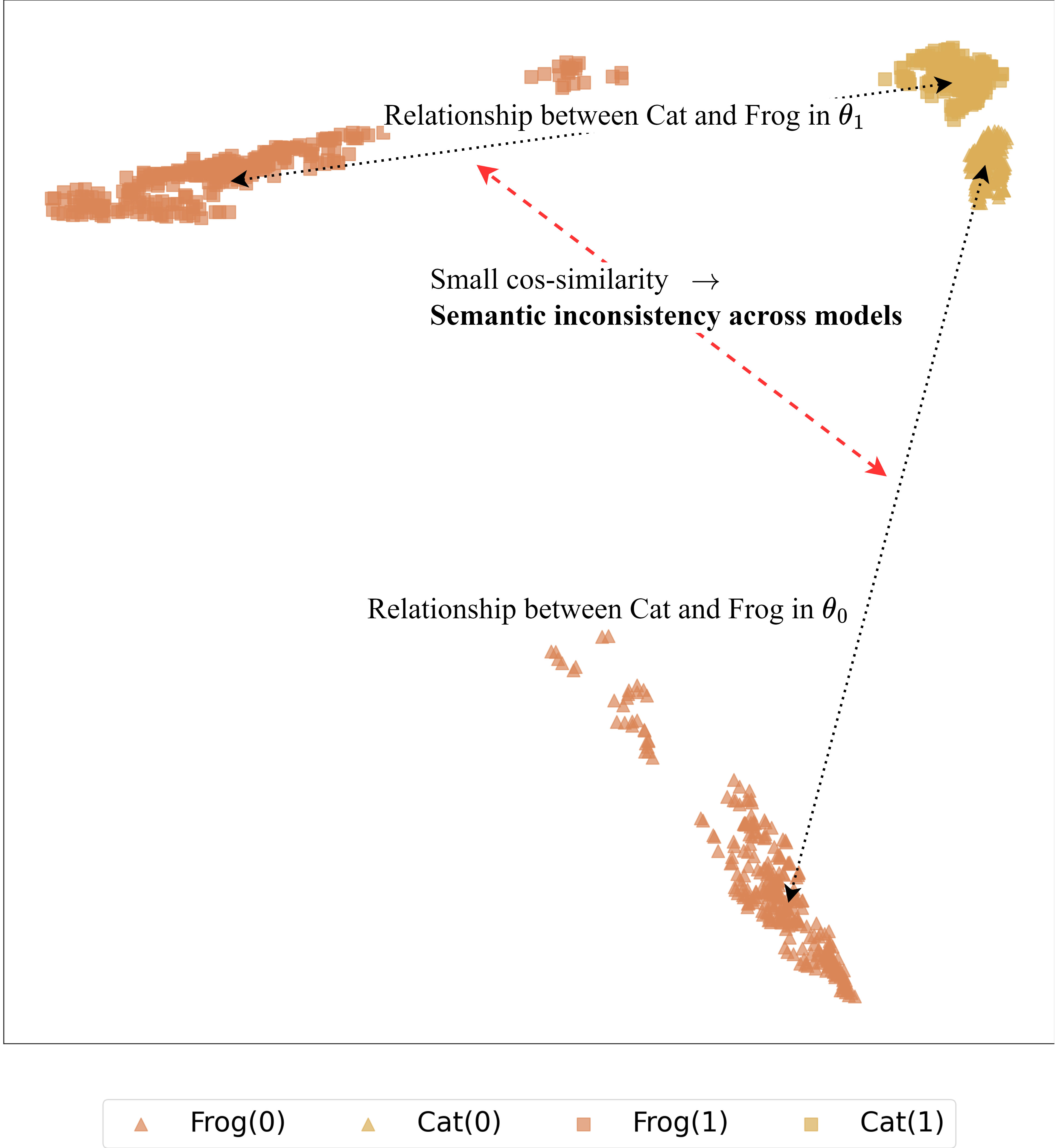}
    \caption{T-SNE results on Semantic inconsistency across models. The models are trained by RoLR on CIFAR-10 with 80\% symmetric noise.}
    \label{fig:siam}
\end{figure}

\cref{fig:siam} demonstrates the t-SNE results on Semantic inconsistency across models.

\subsection{Details of different augmentation strategies}
In our method, we use two augmentation strategies for training.
In particular, the weak augmentation consists of random crop and random horizontal flip. The strong transformation consists of RandAugment \cite{DBLP:conf/nips/CubukZS020}  and Cutout \cite{DBLP:journals/corr/abs-1708-04552}.
RandAugment initially selects a specified number of operations randomly from a predetermined set of transformations, which includes geometric and photometric transformations like affine transformation and color adjustment. Subsequently, these operations are implemented with designated magnitudes. Cutout involves randomly masking square regions of images. These augmentations are then sequentially applied to the input images \cite{Chen2021TwoWD}.


\subsection{Proof of maximizing the lower bound of the mutual information}


For the sake of simplicity in the proof, we exclusively focus on individual positive sample pairs, where the same sample is depicted by different models. Furthermore, based on t-SNE analysis, it is apparent that samples from distinct perspectives can be readily aligned, as illustrated in \cref{fig:rolr_view_1}. Consequently, we do not factor in the distinctions between various views in this proof.
So, in \cref{eq:CCLRL_1}, given the anchor embedding $f_{\theta_m}(x)$ from $\theta_m$ and contrastive embeddings $f_{\theta_{(1-m)}}(x)$ from $\theta_{(1-m)}$, we formulate the $(f_{\theta_m}(x),f_{\theta_{(1-m)}}(x))$ as the positive pair and $\{(f_{\theta_m}(x),f_{\theta_{(1-m)}}(x_j))\}_{j=1}^{N-1}$ as negative pairs. 
To generalize, we set $m=0$ in the proof.
The the joint distribution is $\mu(f_{0},f_{1})$ and the product of marginals is $\mu(f_{0})\mu(f_{1})$.
The distribution $\mathbf{q}$ with an indicator variable $K$ can represent whether a pair $(f_{0},f_{1})$ is drawn from the joint distribution ($K=1$) or the product of marginals ($K=0$):

\begin{equation}
    \begin{aligned}
        \mathbf{q}(f_{0},f_{1}|K=1) &= \mu(f_{0},f_{1}) \\
        \mathbf{q}(f_{0},f_{1}|K=0) &= \mu(f_{0})\mu(f_{1}) 
    \end{aligned}
\end{equation}

Therefore, $K=1$ can also indicate the positive pair $(f_{0}(x),f_{1}(x))$ while $K=0$ can indicate a negative pair from $\{(f_{0}(x),f_{1}(x_j))\}_{j=1}^{N-1}$. 
Based on our approach and the assumptions mentioned earlier, we have one positive pair for every $N-1$ pairs.
Therefore, the prior probabilities of the variable $K$ are:
\begin{equation}
    \begin{aligned}
        \mathbf{q}(K=1) &= \frac{1}{N} \\
        \mathbf{q}(K=0) &= \frac{N-1}{N} 
    \end{aligned}
\end{equation}

We use Bayes' rule to derive the class posterior of the pair $(f_{0},f_{1})$ belonging to the positive case ($K=1$).

\begin{equation}
    \begin{aligned}
        &\mathbf{q}(K=1|f_{0},f_{1}) \\
        &= \frac{\mathbf{q}(f_{0},f_{1}|K=1)\mathbf{q}(K=1)}{\mathbf{q}(f_{0},f_{1}|K=1)\mathbf{q}(K=1) + \mathbf{q}(f_{0},f_{1}|K=0)\mathbf{q}(K=0)}\\
        &=\frac{\mu(f_{0},f_{1})}{\mu(f_{0},f_{1}) + (N-1)\mu(f_{0})\mu(f_{1}) }
    \end{aligned}
\end{equation}

The log class posterior can be further represented as follows:
\begin{equation}
    \begin{aligned}
        &\text{log}\mathbf{q}(K=1|f_{0},f_{1})\\
        &=\text{log}\frac{\mu(f_{0},f_{1})}{\mu(f_{0},f_{1}) + (N-1)\mu(f_{0})\mu(f_{1})}\\
        &=-\text{log}(1+(N-1)\frac{\mu(f_{0})\mu(f_{1})}{\mu(f_{0},f_{1})}) \\
        &\leq -\text{log}(N) + \text{log}\frac{\mu(f_{0},f_{1})}{\mu(f_{0})\mu(f_{1})}
    \end{aligned}
\end{equation}

By calculating expectations over the log class posterior, we can establish a connection to the mutual information in the following manner:
\begin{equation}
    \begin{aligned}
        &\mathds{E}_{\mathbf{q}(f_{0},f_{1}|K=1)}\text{log}\mathbf{q}(K=1|f_{0},f_{1}) \\
        &\leq -\text{log}(N) + \mathds{E}_{\mu(f_{0},f_{1})}\text{log}(\frac{\mu(f_{0},f_{1})}{\mu(f_{0})\mu(f_{1})})\\
        &=-\text{log}(N) + I(f_{0},f_{1})
    \end{aligned}
\end{equation}
In fact, CCLRL can be seen as the negative log class posterior of the positive pair:
\begin{equation}
    \mathcal{L}_{\text{CCLRL}} = -\text{log}\mathbf{q}(K=1|f_{0},f_{1})
\end{equation}
Therefore, we can connect $\mathcal{L}_{\text{CCLRL}}$ to the mutual information $I(f_{0},f_{1})$ as follows:

\begin{equation}
    \begin{aligned}
        &\mathds{E}[\mathcal{L}_{\text{CCLRL}}]\geq \text{log}(N) - I(f_{0},f_{1}) \\
        \Rightarrow &I(f_{0},f_{1}) \geq \text{log}(N) -  \mathds{E}[\mathcal{L}_{\text{CCLRL}}]
    \end{aligned}
\end{equation}

\subsection{Details of the experimental setup}

\subsubsection{Brief introduction of the datasets.}

CIFAR-10 and CIFAR-100 are labeled subsets of the 80 million tiny images dataset, with 50000 training colour images and 10000 test colour images in 10 classes (100 classes). 
CIFAR-10N and CIFAR-100N equip CIFAR-10 and CIFAR-100 with human-annotated real-world noisy labels researchers collected from Amazon Mechanical Turk. They have various types of noise such as Aggregate, Random, Worst or Fine.
Animal-10N \cite{Song2019} is a noisy dataset from the real world, with a noise rate expected to be around 8\% and 50000 training colour images and 5000 test colour images in 10 confusing classes.
WebVision \cite{DBLP:journals/corr/abs-1708-02862} comprises web-crawled images with similar concepts as ImageNet ILSVRC12 \cite{DBLP:conf/cvpr/DengDSLL009}. We follow the previous works \cite{Chen2021TwoWD,DBLP:conf/iccv/Zhang0FLCLL23} and compare baselines on the first 50 classes of ImageNet ILSVRC12 dataset.

\subsubsection{Brief introduction of comparison methods.}

\begin{itemize}
    \item \textit{Decoupling} \cite{Malach2017} sends samples from the two models with inconsistent outputs to each other for training.
    \item \textit{Co-teaching} \cite{Han2018c} trains the models using the selected clean samples and \textit{co-training} mechanism\cite{DBLP:conf/colt/BlumM98}.
    \item \textit{Co-teaching+} \cite{Yu2019a} further develops \textit{Co-teaching} and only used disagreement samples in selected small-loss samples for training.
    \item \textit{JoCoR} \cite{DBLP:conf/cvpr/WeiFC020} uses two networks together to get two different views of the dataset but binds them by an joint loss, making their prediction consistent.
    \item \textit{APL} \cite{DBLP:conf/icml/MaH00E020} assemblages two robust loss functions that mutually boost each other.
    \item \textit{Co-learning} \cite{DBLP:conf/mm/TanXWL21} aligns the knowledge from supervised and unsupervised learning in a cooperative way to ensure that the model learns clean knowledge.
    \item \textit{Cycle-Consistency Reg.} \cite{Cheng2022ClassDependentLL} reduces the side-effects of the inaccurate noisy class posterior through a novel forward-backward cycle-consistency regularization.
    \item  \textit{SELC} \cite{Lu2022SELCSL} refines the model by gradually reducing supervision from noisy labels and increasing supervision from ensemble predictions to retain the reliable knowledge in early stage of training.
    \item \textit{LC} \cite{Wei2022MitigatingMO} clamps the norm of the logit vector to mitigate the overfitting to noisy samples. 
    \item \textit{RankMatch} \cite{DBLP:conf/iccv/Zhang0FLCLL23} propose rank contrastive loss, which strengthens the consistency of similar samples regardless of their potential noisy labels and facilitates feature representation learning. 
    \item \textit{DMLP \cite{DBLP:conf/cvpr/TuZLLLWWZ23}} decouples the label correction process into label-free representation learning and a simple meta label purifier and can be pluged into various LwNL methods.
\end{itemize}

\subsubsection{Implementation Details.}
For the classification task,
we implement all methods with default parameters in Pytorch 1.8, and conduct all the experiments on NVIDIA 3090 GPU. 
We utilize ResNet-18\cite{He_2016_CVPR} for CIFAR and CIFAR-N datasets. For Animal-10N and WebVision, we use ResNet-50 \cite{He_2016_CVPR}.
For the fair comparison, we choose Adam optimizer (momentum=0.9) is with an initial learning rate of 0.001, and the batch size and the epoch are set to 128 and 500 for CIFAR-10, CIFAR-100, CIFAR-10N and CIFAR-100N. 
For the different comparison methods the hyperparameters have been set according to those given in the original paper.
The warm-up epoch is 10 epochs for CIFAR-10 and 30 epochs for CIFAR-100, respectively.
In our method, we set $c=0.95,T = 0.5$ as the default value.
We report the average performance of our method over 3 trials with different random seeds for generating noise and parameters initialization.


\subsection{More Comprehensive Analysis on Semantic Contamination.}
We have incorporated additional metrics and analyses to further elucidate the performance of our method. Specifically, we employ LCA metric \cite{DBLP:conf/icml/ShiGTCLVFFK24}, which quantifies the semantic distance between two classes using class taxonomy. A lower LCA score between the top-1 and top-2 logits indicates that the model has learned a closer semantic relationship, implying that the model experiences reduced levels of semantic contamination (SC). We applied this metric to the model's output to provide a quantitative assessment of our method's performance, as detailed in \cref{tab:app_sc}. The experimental results demonstrate that our method achieves better semantic relevance, providing an experimental basis for the analysis in \cref{fig:intro_1}.

\begin{table}[htbp]
    \centering
    \begin{tabular}{l|c}
    \toprule
    Method & LCA$\downarrow$ \\
    \midrule
    Co-teaching&	4.58\\
    DivideMix& 3.21\\
    RoLR&	3.03\\
    Ours&	\textbf{2.72}\\
    \bottomrule
    \end{tabular}
    \caption{LCA Results on CIFAR-10 with 80\% Symmetric Noise.}
    \label{tab:app_sc}
\end{table}

\subsection{More discussion on Computational Efficiency.}
The computational complexity of our method is $O(M\cdot N+2\cdot B \cdot N)$
, where $M$ and $N$ represent the sizes of the models and dataset, respectively, and $B$ denotes the batch size.

\subsection{Comparison with more SOTA methods on Mini-Web dataset.}
NGC \cite{DBLP:conf/iccv/Wu0JMTL21} and Sel-CL+ \cite{DBLP:conf/cvpr/LiXGL22} are two important baselines for comparison. We supplement the relevant results, as shown in \cref{tab:app_miniweb}. The experimental results show that our algorithm achieves better performance and enhances the robustness of the model.
\begin{table}[htbp]
    \centering
    \begin{tabular}{l|cc}
    \toprule
    Method & top1 & top5 \\
    \midrule
    NGC &	79.2 &	91.8\\
 Sel-CL+	& 80.0 &	92.6\\
 Ours&	\textbf{82.3}&	\textbf{94.6}\\
    \bottomrule
    \end{tabular}
    \caption{More results on Mini-Web Dataset.}
    \label{tab:app_miniweb}
\end{table}

\end{document}